\documentclass{article}

     \usepackage[nonatbib, preprint]{neurips_2020}

\usepackage[utf8]{inputenc} 
\usepackage[T1]{fontenc}    
\usepackage{hyperref}       
\usepackage{url}            
\usepackage{booktabs}       
\usepackage{amsfonts}       
\usepackage{nicefrac}       
\usepackage{microtype}      
\usepackage{comment}

\title{A Meta-Learned Neuron model for Continual Learning}

\author{%
  Rodrigue Siry \\
  Université de Caen Normandie - GREYC - CNRS\\
  \texttt{rodrigue.siry@unicaen.fr} \\
}

\begin{document}

\maketitle

\begin{abstract}
Continual learning is the ability to acquire new knowledge without forgetting the previously learned one, assuming no further access to past training data. Neural network approximators trained with gradient descent are known to fail in this setting as they must learn from a stream of data-points sampled from a stationary distribution to converge. In this work, we replace the standard neuron by a meta-learned neuron model whom inference and update rules are optimized to minimize catastrophic interference. Our approach can memorize dataset-length sequences of training samples, and its learning capabilities generalize to any domain. Unlike previous continual learning methods, our method does not make any assumption about how tasks are constructed, delivered and how they relate to each other: it simply absorbs and retains training samples one by one, whether the stream of input data is time-correlated or not.

\end{abstract}

\section{Introduction}
The past decade saw an unprecedented burst in deep-learning based approaches. Consistent discovery of new neural architectures and training objectives eventually had them overtaking state-of-the-art on nearly any machine learning problem; achieving impressive feats in sample classification, generative modeling and reinforcement learning. The common ingredient of all these methods is neural network approximators trained by gradient descent. Ability of gradient descent training to find a good solution to any differentiable objective-fitting problem is not to be proved anymore. However, it needs access to an i.i.d. sampler of datapoints during the whole training. This put uncomfortable  constraints on how the lifetime of a learning agent can be managed: first, a training phase with full access on training data, second, an inference phase where the trained agent is frozen and used. Such constraints make current AI models unsuited to the real world, where new skills must be learned continuously without affecting previous knowledge, and where training data is only available at a time. In this paper, we replace the standard neuron model trained by gradient descent by a more sophisticated parametric model, whom local inference and training dynamics are jointly optimized so that an assembly of such neurons displays continual learning as an emergent property.

\section{Definitions and Related work}
\subsection{Continual learning}

Let $\mathcal{D} = \{(x_i, y_i)\}_{1<i<T}$ be a sequence of training samples of $\mathcal{X}\times\mathcal{Y}$, ordered by the time they will be presented to the learning agent. For commodity, we use the compact notation $\mathcal{D}_{i<t}$ for $\{(x_i,y_i)\}_{i<t}$.\\
Continual learning is the ability to learn from new datapoints $(x_t, y_t)$ while protecting the knowledge acquired from the previously learned ones (the set $\mathcal{D}_{i<t}$), assuming no further access to $\mathcal{D}_{i<t}$. To assess the difficulty of the continual learning problem, studying the time-correlations found in the training sequence is crucial: intuitively, the more time-correlated the sequence, the harder. It is important to note that most previous contributions of the continual learning literature make strong assumptions about these time-correlations. In the \textit{task-incremental} case, for example, we assume that the input sequences can be decomposed in a sequence of "tasks" $\mathcal{T}_i$. Tasks are usually built from a specific modality of the whole input dataset (for example a subset of all classes) and task boundaries are known during learning. The \textit{class-incremental} case is the specific case in which each task is composed of samples from a single new class.

We now proceed to an overview of existing continual learning methods. Gradient descent, the most popular learning algorithm, is prone to \textit{catastrophic interference}: following the gradient computed from recent samples quickly deteriorates the knowledge of the model that has been acquired on past samples. Several methods were hence proposed to mitigate forgetting:

Replay methods \cite{icarl} select and store a small subset of past samples that best enforce the remembering constraint when replayed. However, the memory buffer must grow linearly with the number of tasks. Surprisingly, maintaining a fixed-size but class-balanced buffer, and then train from scratch provides a good baseline that does not rely on the task-incremental hypothesis \cite{gdumb}. Architecture growing methods \cite{architecture_1, architecture_2} add capacity to the model when a new task arrives while freezing weights related to previous tasks. This, again, do not scale well with the number of tasks and is limited to the task-incremental case.\\
EWC \cite{ewc} provides an elegant bayesian formulation of the task-incremental learning problem, and uses an approximate posterior distribution of the network weights after observing $\{\mathcal{T}_i\}_{i\leq t}$. In practice, it evaluates the importance of each parameter value at solving prior tasks, and lowers the plasticity of important parameters with a soft constraint when learning a new task. While giving decent results, EWC is known to become more conservative as the number of tasks becomes too large, limiting its ability to learn, and is limited to the task-incremental case. \\
Meta-learning methods \cite{oml, anml} brought a new take on continual learning and pushed state-of-art further: instead of using hand-made heuristics, they directly optimize the objective of learning continually w.r.t. some meta-parameter. The drawback is that those methods are data-driven: they must be trained on a set of \textit{meta-train} tasks and can hence be prone to overfitting, compromising their ability to generalize to new, \textit{meta-test} tasks. Careful design and regularization of the meta-parameter is hence crucial to achieve meta-generalization. OML \cite{oml} proposes to meta-learn an encoder representation that facilitates continual learning for a downstream classifier. ANML \cite{anml} improves on OML by adding a neuro-modulatory network whose role is to implicitly protect parts of the base network from catastrophic interference by modulating activations. In both cases, we meta-learn the initialization of a large neural network function: the meta-parameter is therefore \textit{global} and high-capacity. This raises concern about the ability of such methods to generalize to new domains \cite{meta_overfitting}. So far, OML and ANML were only tested on tasks built with the Omniglot dataset \cite{omniglot} in which all classes largely share a lot of common features, hence facilitating the transfer from meta-train to meta-test tasks.

\subsection{Meta update rule}
In this paper, we address continual learning with a meta-learnt neuron model. Meta-learnt learning rules of various complexity were already leveraged for faster i.i.d. supervised learning \cite{meta_faster_learning} and unsupervised learning \cite{meta_repr_learning}. Contrary to representations, update rules are constrained, low-capacity models. We expect them to discover an emergent learning algorithm that enables continual learning of the base network. For this reason, we expect such a strategy to generalize to train sequences containing unseen domains and classes. Additionally, we can meta-train it so it can handle any degree of time-correlation in the input sequence.

\section{The NEM model}
\subsection{Introduction}
Unlike most previous works from the literature \cite{ewc, icarl, architecture_1, architecture_2}, we address continual learning of a stream of samples instead of a stream of tasks. This approach is more challenging as it forbids any conditioning by task label and requires an inner update of the model for each training sample. However, it encompasses all sub-cases of continual learning: in a sequence of samples, all sorts of time-correlation can be imagined, spanning from the class-incremental case (time and label are correlated maximally) to the i.i.d. case (input samples are shuffled, no input statistic correlates with time), complying with how learning should occur in the real world where the notion of task is not well-defined.\\
Like \cite{oml, anml}, we consider meta-learning as a promising avenue to solve continual learning. However, meta-learning comes with the risk of meta-overfitting \cite{meta_overfitting}: the more capacity a meta-parameter has, the higher the risk for the model to specialize on meta-train tasks and failing on meta-test tasks. This is a known problem of MAML \cite{maml}, where we meta-learn the initial parameter $\theta_{init}$ of a classifier network. With such a meta-parametrization, it is easy to imagine a degenerate solution to the meta-learning objective in which $\theta_{init}$ produces representations that are meaningful to meta-training classes but do not transfer to meta-test classes at all \cite{meta_overfitting}. The continual learning models proposed in \cite{oml, anml} should suffer the same burden: their meta-parametrization being \textit{global}, we expect them to overfit to the meta-train domain at some point. In this paper, we instead design a parametric, neuron-local update rule shared among all neurons of the base network. We then optimize this update rule so that its emerging behavior enables the base network to continuously absorb new datapoints while protecting its behavior on previous ones, hence building \textbf{Neurons for Emergent Memorization (NEM)}. As we learn a local, compact \textit{learning rule} instead of a global, high capacity \textit{initial representation}, we expect our model to work on any other meta-test domain.

\subsection{Description of the NEM unit:} In this paragraph, we proceed to describe the parametrization of our meta-learned NEM neuron unit, which is a generalization of the standard deep learning neuron. 
Generally speaking, a neuron can be seen as a tiny dynamical system able to exchange messages and adjust connections with its direct neighbors. Wired assemblies of neurons are expected to guarantee efficient inference and lifelong learning of the whole network as an emergent property. We designed the parametrization of NEM so that it fits this statement, without being more specific.

Our parametrization improves upon the standard neuron on two aspects: 1) neurons exchange \textit{vector} activations instead of scalars ones and 2) each individual neuron maintains a hidden state vector, allowing long-term dynamics to be learned within each unit.\\
Inference and learning in NEM networks globally resembles traditional forward pass/gradient backpropagation: when a sample $x_i$ is fed into the network, each NEM unit receives a \textit{forward message}, which is the weighted sum of previous layer activations. It then outputs a forward vector activation that will be used by the next layer, and so on until the last layer is attained. The vector activation from each output neuron can then be collapsed into a scalar classification logit. If we want the network to learn $(x_i, y_i)$, we continue and provide the information about $y_i$ in the form of vector \textit{backward messages} that are sent to each one of the output neurons. They use it to update their internal state, then produce additional backward messages that will be sent to the previous layer and so on until the first neurons are reached. Finally, when all neurons have received both their forward and backward message, they use this context information to update their wiring, so the base network fits $(x_i, y_i)$.

\textbf{Forward pass:} 
Similar to standard neurons, the NEM unit is composed of $N$ inputs and $1$ output. It receives $N$ activations from the previous layer and produces a single activation at output. To allow neurons to exchange complex messages, we replace the standard \textit{scalar} activations by \textit{vectors} of fixed dimension $d_a$. The input weights, however, remain scalar. Furthermore, we allow each neuron to maintain an hidden state vector $h$ of dimension $d_h$. To produce the output $s$, we first compute the forward message $z = \sum\limits_N w_i x_i \in \mathbb{R}^{d_a}$, which is the analog of pre-activations in standard neural network. We then combine the forward message and the hidden state with a learned activation function $f_{act}: \mathbb{R}^{d_a+d_h}\rightarrow \mathbb{R}^{d_a}$. 
\begin{center}
$s =f_{act}\left(z, h\right)$  
\end{center}

The combination of hidden state $h$ and learned $f_{act}$ offers a flexible neuron model: the information stored in $h$ can be either used to influence forward inference (ex: generalize the bias) or store long-term, hidden information that might be useful for lifelong knowledge consolidation.\\
To convert the input datapoint $x_i$ into the set of vector activations fed to the first layer of NEM units, we apply a learned expansion function $f_{expand}:\mathbb{R}\rightarrow\mathbb{R}^{d_a}$ dimension-wisely. Similarly, to obtain scalar classification logits from last layer activations, we learn $f_{collapse}:\mathbb{R}^{d_a}\rightarrow\mathbb{R}$.

\textbf{Backward pass and update Rule:} We now proceed to describe how learning occurs in a network of NEM units.
Let $\mathcal{D} = \{(x_i, y_i)\}_{1<i<T}$ be a sequence of training samples of $\mathcal{X}\times\mathcal{Y}$.
The goal of the update rule is to modify the input weights $\{w_i\}$ and hidden state $h$ of each neuron so that upon reception of a sample $(x_t, y_t)$, the network learns $(x_t, y_t)$ while protecting its behavior on $\mathcal{D}_{i<t}$.\\
We first perform a full forward pass over the network and keep the pre-activation $z$ of every neuron. To inject the label information $y_i$ and bootstrap the backward pass, we then provide to every neuron from the last layer a \textit{backward message} $d\in \mathbb{R}^{d_a}$. All dimensions of $d$ are set to '$1$' for the output neuron concerned by the right label and $d=0$ for the others neurons. Upon reception of its backward message, a NEM unit updates its internal state by using the contextual information from $z$ and $d$. This update is parametrized by a learned $f_{update}:\mathbb{R}^{d_a+d_a+d_h}\rightarrow\mathbb{R}^{d_h}$

\begin{center}
$h_{t+1} =f_{update}\left(z, d, h_t\right)$  
\end{center}
To back-propagate the learning signal to the previous layer, we define the backward message received by the $i^{th}$ neuron from layer $l-1$ to be the weighted sum of all hidden states of neurons of layer $l$. The weight matrix being the transpose of the forward weight matrix. 
\begin{center}
$d^{l-1}_i = \sum\limits_{j} w_{ij} h^l_{j, t+1}$
\end{center}

Once the backward pass is achieved and all internal states have been updated, we may finally update the weights connecting NEM units. In a layer $l$, the update of weight $w^l_{ij}$ only depends on $h^{l-1}_{i, t+1}$ and $h^l_{j, t+1}$. We learn projections $f_{prev}:\mathbb{R}^{d_h}\rightarrow\mathbb{R}^{d_h}$, $f_{next}:\mathbb{R}^{d_h}\rightarrow\mathbb{R}^{d_h}$ so that:

\begin{center}
$w^l_{ij, t+1} = w^l_{ij, t} + \gamma\cos(f_{prev}(h^{l-1}_{i, t+1}), f_{next}(h^l_{j, t+1}))$
\end{center}
Where $\gamma$ is a small learning rate and $\cos(x,y)$ the cosine distance between two vectors.
\subsection{Meta-training objective and optimization}
Let $\mathcal{D} = \{(x_i, y_i)\}_{1<i<T}$ be a sequence of training samples of $\mathcal{X}\times\mathcal{Y}$.
To catch up with the formalism of meta-learning, we note $\Psi$ the meta-parameter: the union of the parameters of all functions ($f_{act}, f_{update}, f_{prev}, f_{next}, f_{expand}, f_{collapse}$) defining the dynamics of the NEM update rule. We note $\theta$ the base parameter: the union of base network edges weights and neurons inner states. Our update rule hence defines the transition function $T_\Psi$:
\begin{center}
    $\theta_{t+1}= T_{\Psi}(\theta_t, (x_t, y_t))$
\end{center}

We build a continual learning task by sampling an input sequence $\mathcal{D} = \{(x_t, y_t)\}_{1\leq i\leq T}$. To learn an update rule that is robust to all learning regimes, we sample sequences with varying degrees of time-correlation, spanning from the class-incremental case to the random shuffle case.

We start the continual learning task by sampling random initial edge weights and zero inner states for all NEM units in the network.
We then unroll the whole sequence absorption process by feeding the samples $(x_t, y_t)$ one at a time to the network and apply the update rule. At time $t$, we want the network to classify any past sample from $\mathcal{D}_{i<t}$ correctly. Once the sequence has been entirely fed to the model, we evaluate the objective, backpropagate through the whole recurrent graph and perform a meta-train step to optimize the NEM update rule: \\
\begin{center}
$\min\limits_\Psi\sum\limits_{1 \leq t \leq T}\mathcal{L}(\theta_t, 
\mathcal{D}_{i\leq t})$.\\
\end{center}

In practice, optimization of complex recurrent systems across many timesteps is slow, memory-intensive and unstable. Additionally, our problem inherently requires memorization of arbitrarily long range dependencies. Several design and training choices were hence crucial to ensure stable convergence and regularization of NEM:\\
\textbf{Progressive increase of sequence length}: directly meta-optimize on long input sequences is likely to never converge. We found that using curriculum learning was crucial to unlock optimization: we start by training the model to memorize sequences of length 2, then 10, 100 and so on by gradually increasing sequence length. \textbf{Small base edge weight update}: similarly to \cite{meta_repr_learning}, we found mandatory to constrain the update rule to perform small updates of the base network wiring. Doing otherwise produces chaotic meta gradients and stale learning. We hence set $\gamma$ to a small value (ex: $10^{-4}$). \textbf{Large activation and hidden vector dimension}: setting $d_h$ and $d_a$ to a relatively large value (256) facilitates optimization. \textbf{Truncated BPTT}: Even with a small base model, unrolling the whole episode for more than 1000 updates was not feasible with our hardware. We therefore used truncated back-propagation through time to save GPU memory at the cost of biasing the meta-gradient estimate. \textbf{Outer-loop drop-out}: With all the tricks presented so far, the meta-training of $\Psi$ converges to a solution that makes the base model good at memorizing training samples. However, at this point, we found that the generalization capabilities of the base network of NEM units were far worse than a model trained by gradient descent from an equivalent number of memorized training samples. The natural solution to this problem would be to compute the outer-loop loss with a test set that is distinct from the training sequence in order to meta-learn generalization. Unfortunately, we observed that minimizing this criterion leads to stale optimization. We therefore instead define a proxy regularization inspired from dropout\cite{dropout}: when evaluating $\mathcal{L}(\theta_t, 
\mathcal{D}_{i\leq t})$ in the outer-loop criterion for some $t$, we drop $10\%$ of the input activations of a randomly chosen layer. However, we do not apply this dropout during the forward pass required for each inner-loop update. This regularization do not hinder meta-optimization, avoids pathological co-adaptation of NEM units and rectifies the generalization capabilities of the base NEM network to a level only slightly below gradient descent.

\section{Experiments}

\textbf{Meta-train tasks:} We generate a meta-training input sequence from 32x32 image samples of the tiny-ImageNet-200 dataset (200 classes). To further reduce the input dimension, we convert all images to grayscale, leading to 1024-dim flat input vectors. When generating a sequence of given length, we randomize the sample/class ratio. For example, a sequence of length 100 can be composed of \{2 classes; 50 samples/class\}, \{100 classes; 1 sample/class\} or every other intermediate ratio. Sequences are initially built class-incremental, but to allow our learning rule to handle diverse time-correlation regimes in the input stream, we shuffle the sequence with probability $p=0.5$.

\textbf{Meta-test tasks:} To assess the domain-generalization capability of our meta-trained update rule, we first construct meta-test sequences with gray-scaled 32x32 images from the CIFAR-100 dataset (100 classes). We vary the number of classes and evaluate both the class-incremetal scenario (NEM CI) and the shuffled sequence scenario (NEM shuffle). To further assess the domain-universality of the NEM update rule, we test on additional 10-class sequences built from the MNIST\cite{mnist} and SVHN\cite{svhn} datasets. 

\textbf{Base architecture:} The base model is a feedforward architecture of size $1024\rightarrow 1024\rightarrow 1024 \rightarrow 200$ composed of meta-learned NEM units, totaling $2248$ neurons. We set the hidden state and output dimensions to $d_h = d_a = 256$. In all other baselines, we keep the same architecture but replace the NEM units by standard neurons with ReLU activation. Note that this architecture does not benefit from inductive bias like convolutional neural networks, it is therefore not surprising to observe relatively low test accuracies in the experiments.

\textbf{Meta architecture:} To make the NEM model as lightweight as possible, we implement $f_{prev}, f_{next}, f_{expand}$ and $f_{collapse}$ as simple linear projections. To introduce non-linearities in the base network, we implement $f_{act}$ as a linear function followed by a ReLU activation. Finally, we implement $f_{update}$ as a GRU\cite{gru} recurrent neural network, whose hidden state is the neuron hidden state.

\textbf{Training details and hyperparameters} We meta-optimize NEM by following a curriculum of increasing sequence lengths $[2, 10, 50, 100, 500, 1k, 2k, 5k, 10k]$. The next level of difficulty is unlocked when the model reaches an average memorization accuracy of $0.9$. To save compute and memory, we compute a monte-carlo estimate of the meta-objective from a single sequence, by evaluating $\theta_t$ on a single sample from $\mathcal{D}_{i<t}$ at each timestep $t$. We use Adam as outer-loop optimizer, with an initial learning rate of $1e^{-4}$ that is then lowered to $1e^{-5}$ for sequences longer than $1k$ samples. TBPTT is performed with truncation size $100$ so that the unrolled segment fits in a 1080-Ti GPU.

\textbf{Choosing appropriate baselines:} Continual learning methods usually need to store extra information about the past in addition to the base model parameter. It is hence crucial to measure how this memory overhead grows w.r.t. input sequence length and only compare methods with similar memory budgets. This information is explicit in buffer-based methods, as it is simply the size of the replay buffer. In our case, the memory overhead is the total size of all hidden states, it weights as much as a buffer of $\frac{2248 \times 256}{1024} = 562$ images.
\begin{itemize}
    \item \textbf{GDumb}: GDumb \cite{gdumb} is a very simple baseline that consists in first building a class-balanced buffer as the samples arrive, then naïvely train the model on the buffer. Surprisingly, GDumb provides competitive performance that put into question the usefulness of most previous works on continual learning. A buffer of size $600$ roughly corresponds to the memory footprint of NEM.
    
    \item \textbf{Full-iid}: To get an upper-bound on model test accuracy, we perform a standard batch gradient-descent training with full access to the dataset.
\end{itemize}

We test NEM on a wide variety of test sequences. In tables \ref{tab:cifar100}-\ref{tab:otherdomains}, we report both memorization (train set) and generalization (test set) accuracies, averaged over 10 random seeds. 

\begin{table}[htb]
    \centering
    \hfill
    \begin{tabular}{*3c}
         \multicolumn{3}{c}{CIFAR-100: 10 classes, 500 samples / class}\\
         \toprule
          Method &  train& test\\
         \midrule
         GDumb-600& 0.39& 0.31\\
         NEM (CI)& 0.91 & 0.38 \\
         NEM (shuffled)& 0.91 & 0.38 \\
         Full-iid & 0.99& 0.41\\
         \bottomrule
    \end{tabular}
    \hfill
    \begin{tabular}{*3c}
         \multicolumn{3}{c}{CIFAR-100: 50 classes, 200 samples / class}\\
         \toprule
          Method &  train& test\\
         \midrule
         GDumb-600 & 0.13 & 0.08\\
         NEM (CI) & 0.85 & 0.15 \\
         NEM (shuffled) & 0.84 & 0.15\\
         Full-iid & 0.99 & 0.17\\
         \bottomrule
    \end{tabular}
    \hfill\null
    \caption{Train/Test sets accuracies on CIFAR-100 sequences, with two different class/sample ratios. We either keep the sequences class-incremental (CI) or shuffle them (shuffled), note that baselines are unaffected by this choice, all experiments are averaged over 10 seeds}
    \label{tab:cifar100}
\end{table}

\begin{table}[htb]
    \centering
    \hfill
    \begin{tabular}{*3c}
         \multicolumn{3}{c}{MNIST: 10 classes, 1k samples / class}\\
         \toprule
          Method &  train& test\\
         \midrule
         GDumb-600& 0.89 & 0.87\\
         NEM (CI)& 0.94 & 0.90\\
         NEM (shuffled)& 0.94& 0.90 \\
         Full-iid & 1.0 & 0.95\\
         \bottomrule
    \end{tabular}
    \hfill
    \begin{tabular}{*3c}
         \multicolumn{3}{c}{SVHN: 10 classes, 1k samples / class}\\
         \toprule
          Method &  train& test\\
         \midrule
         GDumb-600 & 0.45 & 0.39\\
         NEM (CI) & 0.65 & 0.36\\
         NEM (shuffled) & 0.66 & 0.36\\
         Full-iid &  1.0 & 0.61\\
         \bottomrule
    \end{tabular}
    \hfill\null
    \caption{Train/Test sets accuracies on MNIST and SVHN sequences of length 10k, averaged over 10 seeds}
    \label{tab:otherdomains}
\end{table}
In the experiments reported in table \ref{tab:cifar100}, we observe that on the meta-test domain CIFAR-100, our assembly of NEM units outperforms GDumb-600 in terms of sample memorization and test performance with a comparable memory budget and architecture. The performance gap widens as the number of classes increases. In all experiments, NEM performs similarly in the class incremental (CI) case and shuffled case, hinting that the update rule is not affected by the degree of time-correlation of the input sequence. The overall number of classes, however, has an slight impact on memorization. Experiments reported in table \ref{tab:otherdomains} demonstrate that the NEM update rule suffers little from meta-overfitting and still achieves satisfying memorization capabilities on domains very distant from tiny-ImageNet-200, such as MNIST or SVHN. However, it appears that in some cases, NEM is not as good as gradient descent at generalizing, to the point it gets beaten by GDumb-600 on the SVHN problem.
\section{Conclusion}
We presented a meta-learnt neuron model to solve the continual learning problem. Our method, labeled NEM,  is extremely flexible and has several key advantages when compared to other CL methods: it does not make any assumptions about the incoming stream of data, works in all regimes of time-correlation, spanning from class-incremental to random shuffle and has learning capabilities that generalize to any domain. 

Though currently limited to feedforward base models and sequences of maximum size $10k$, NEM displays preliminary, but promising results. In this work, the model and training procedure were scaled to fit the memory limitations of a 1080-Ti GPU. We strongly believe that our meta-optimization is stable enough to be carried on sequences much longer than $10k$ if we increase compute and memory by an order of magnitude. Doing so would make the model able to absorb dataset-length sequences and increase generalization, paving the way to flexible, lifelong AI models. We also hope that our model will open the way to exciting future researches: variants of NEM suited to other base architectures, continual learning of unsupervised objectives or reinforcement learning to name a few.

\bibliographystyle{IEEEtran}
\bibliography{References}
\end{document}